\documentclass[paper]{ieice}
\usepackage[pdftex]{graphicx,xcolor}
\usepackage[fleqn]{amsmath}
\usepackage{cite}
\usepackage{hyperref}
\usepackage{times} 
\usepackage{indentfirst} 
\usepackage{tabularx}
\newcolumntype{Y}{>{\centering\arraybackslash}X}
\usepackage{amssymb}
\field{}

\title{3DKeyAD: High-Resolution 3D Point Cloud Anomaly Detection via Keypoint-Guided Point Clustering}
\authorlist{%
\authorentry{Zi Wang}{}{niigata}\MembershipNumber{}
\authorentry{Katsuya Hotta}{}{iwate}\MembershipNumber{}
\authorentry{Koichiro Kamide}{}{toyama}\MembershipNumber{}
\authorentry{Yawen Zou}{}{toyama}\MembershipNumber{}
\authorentry{Chao Zhang}{}{toyama}\MembershipNumber{}
\authorentry{Jun Yu}{}{niigata}\MembershipNumber{}
}
\affiliate[niigata]{Niigata University}
\affiliate[toyama]{University of Toyama}
\affiliate[iwate]{Iwate University}

\received{2024}{1}{1}
\revised{2024}{1}{1}

\begin{document}
\maketitle
\begin{summary}
High-resolution 3D point clouds are highly effective for detecting subtle structural anomalies in industrial inspection. However, their dense and irregular nature imposes significant challenges, including high computational cost, sensitivity to spatial misalignment, and difficulty in capturing localized structural differences. This paper introduces a registration-based anomaly detection framework that combines multi-prototype alignment with cluster-wise discrepancy analysis to enable precise 3D anomaly localization. Specifically, each test sample is first registered to multiple normal prototypes to enable direct structural comparison. To evaluate anomalies at a local level, clustering is performed over the point cloud, and similarity is computed between features from the test sample and the prototypes within each cluster. Rather than selecting cluster centroids randomly, a keypoint-guided strategy is employed, where geometrically informative points are chosen as centroids. This ensures that clusters are centered on feature-rich regions, enabling more meaningful and stable distance-based comparisons. Extensive experiments on the Real3D-AD benchmark demonstrate that the proposed method achieves state-of-the-art performance in both object-level and point-level anomaly detection, even using only raw features.

\end{summary}

\section{Introduction}

\begin{figure}[tb] 
\centering 
\includegraphics[width=1.0\linewidth]{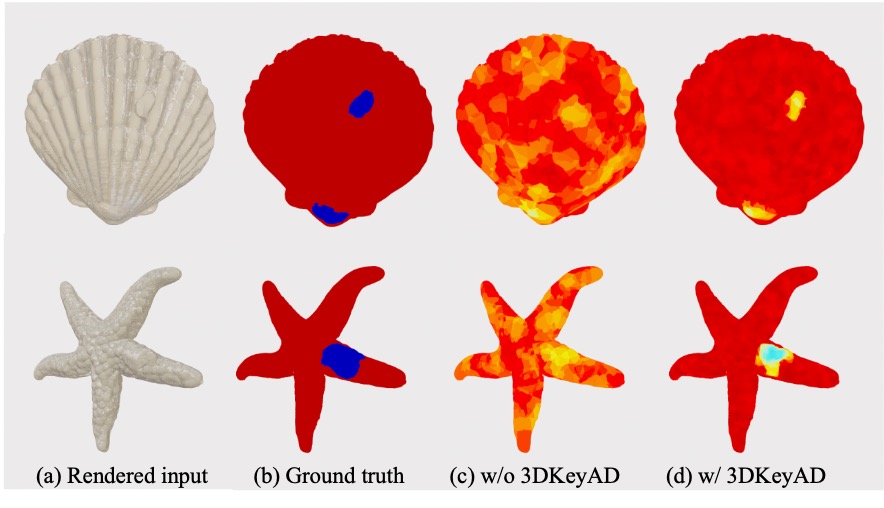} 
\caption{Examples of anomaly detection on high-resolution 3D point clouds. The blue points in the ground truth indicate anomalous regions. After incorporating the keypoint-guided point clustering, our method clearly produces more concentrated anomaly predictions in the true anomalous areas while reducing false positive points.}
\label{fig1} 
\end{figure}

\begin{figure}[tb] 
\centering 
\includegraphics[width=1.0\linewidth]{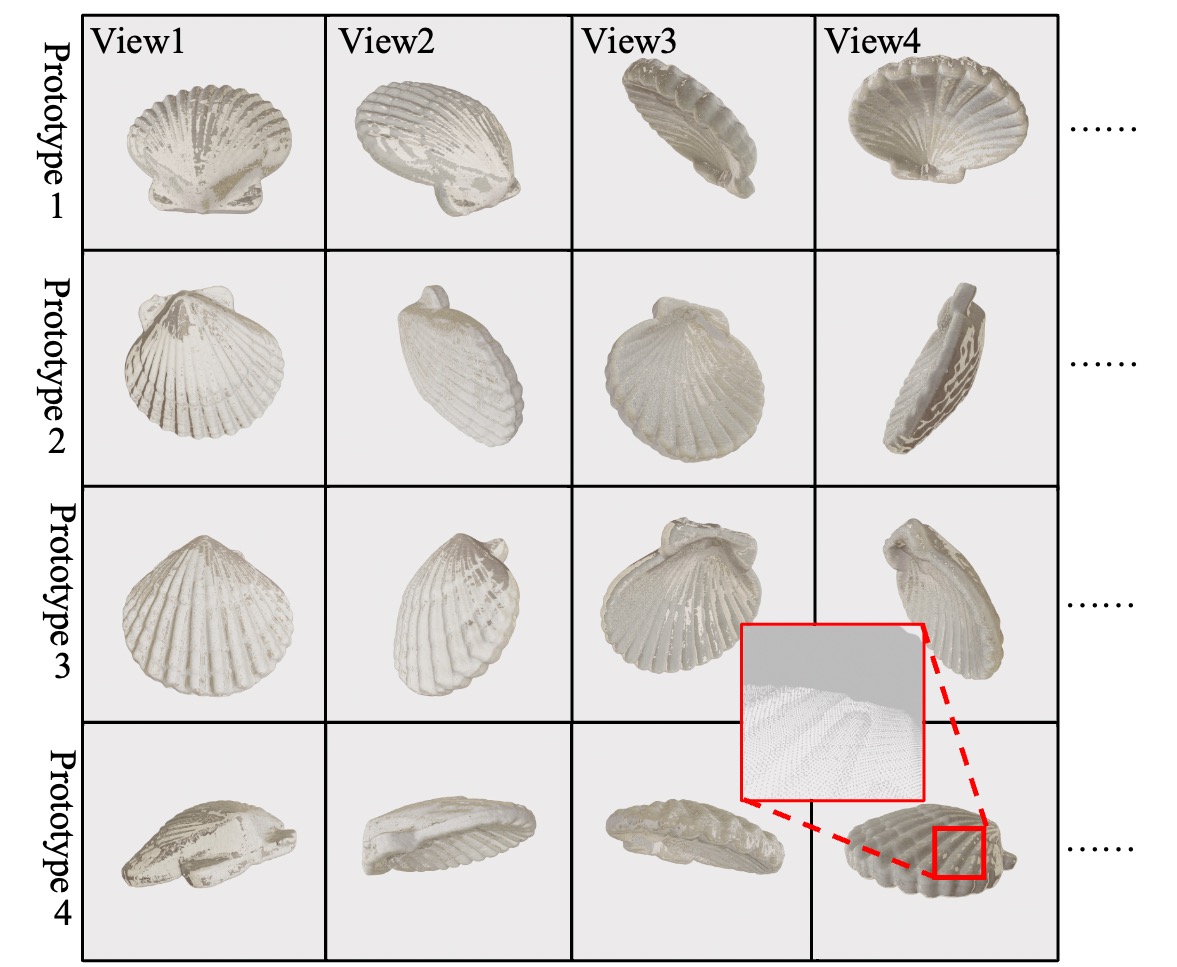} 
\caption{Four prototypes of the \textit{shell} in Real3D-AD dataset.}
\label{fig2} 
\end{figure}

Anomaly detection is a fundamental task in industrial quality control, where the goal is to identify subtle defects or irregularities that deviate from expected norms. While 2D image-based anomaly detection \cite{xie2024iad} has achieved remarkable progress in recent years, particularly with the introduction of large-scale datasets such as MVTec AD \cite{bergmann2019mvtec}, these methods are limited in their ability to capture 3D structural variations. Factors such as lighting conditions, viewpoint changes, and texture ambiguity can obscure surface defects in 2D images. In contrast, 3D point cloud-based anomaly detection provides richer geometric representations that allow for more robust identification of shape-related defects, even in the absence of appearance cues, as illustrated in Fig. \ref{fig1}. With the growing availability of high-resolution 3D sensors and the increasing demand for precise surface inspection, recent research has shifted toward high-resolution 3D anomaly detection. This task differs from conventional settings such as those in MVTec 3D-AD \cite{bergmann2021mvtec}, which are based on depth scans. High-resolution 3D anomaly detection operates directly on dense point clouds containing millions of points per object. In this paper, we use Real3D-AD \cite{liu2023real3d} benchmark, which is specifically designed for high-resolution 3D anomaly detection. As illustrated in Fig. \ref{fig2}, each object in this dataset is represented as a dense point cloud containing thousands to millions of points, providing detailed geometric information suitable for fine-grained inspection. Importantly, the dataset contains only 3D coordinates without color or texture features, making it a purely geometric anomaly detection task. For each object category, four clean reference samples are provided as normal prototypes, while the test set includes both normal and anomalous instances with various types of structural defects. While this allows for fine-grained structural analysis, it also introduces several major challenges. First, the increased point density leads to significantly higher computational demands. Second, high-resolution scans exhibit more spatial variation, making them highly sensitive to misalignment during comparison. Third, the preservation of meaningful local context becomes more difficult, especially when excessive downsampling is required to reduce memory usage. These challenges limit the effectiveness of existing 3D anomaly detection techniques when applied in high-resolution settings.

Many 3D anomaly detection methods follow reconstruction-based designs, such as voxel autoencoders or completion networks, which learn to reconstruct normal shapes and detect anomalies by comparing the output with the input. However, they often fail to capture subtle defects, as the model may incorrectly reconstruct anomalous regions as normal. Other approaches, like Reg3D-AD \cite{liu2023real3d}, use memory banks to store features from normal data and match them against test samples. These methods rely on global matching, which ignores spatial context and can misclassify anomalies that resemble normal regions elsewhere on the object. Some recent methods adopt a teacher–student framework, where the student mimics the teacher’s output on normal data, and discrepancies signal anomalies. While effective for detecting out-of-distribution objects, these models often operate in low-dimensional spaces and lack fine spatial detail, making them less capable of identifying subtle, localized anomalies within a single object.

Besides the above categories, registration-based methods align test samples with normal references in a shared coordinate system. However, they often suffer from the lack of a well-constrained comparison scope. In many cases, anomaly scoring is either computed by comparing each test point to all points in the registered prototypes or restricted to a fixed local neighborhood. Both strategies have significant drawbacks. Global comparison across the entire prototype surface leads to excessive computational cost, which becomes prohibitive for high-resolution point clouds. On the other hand, limiting the comparison to a small fixed neighborhood imposes a strong assumption that the normal samples must exhibit highly consistent local geometry. This assumption is often unrealistic, especially for industrial objects with complex and diverse surface structures. Consequently, either strategy may result in suboptimal detection accuracy, with increased risk of false positives or missed anomalies. A more effective solution requires a structured mechanism to define semantically meaningful comparison regions that balance efficiency and robustness.

To overcome the limitations of prior approaches, we propose a registration-based anomaly detection framework that introduces both structural alignment and localized comparison. Specifically, each test sample is first aligned with a set of normal prototypes through rigid registration, ensuring geometric correspondence in a shared coordinate space. To constrain the comparison scope, we further introduce a keypoint-guided clustering strategy. Rather than comparing each test point to all prototype points or relying on a fixed-radius neighborhood, we detect geometrically salient keypoints and use them as cluster centroids. This allows us to form local regions around structurally meaningful areas, where feature-level discrepancies can be computed more reliably.

By comparing features within clusters centered on keypoints, our method balances efficiency and accuracy. It avoids the high cost of comparing every point and the strict constraint of rigid matching. This makes it effective for detecting fine-grained anomalies on objects with varied shapes. The main contributions of this work are:
\begin{itemize}
    \item We present a novel multi-prototype based anomaly detection framework for high-resolution 3D point clouds, which supports both object-level detection and point-level localization.
    \item We introduce a keypoint-guided clustering mechanism that defines meaningful and flexible comparison regions, improving the sensitivity and robustness of local anomaly scoring.
    \item We demonstrate through extensive experiments on the Real3D-AD benchmark that our method surpasses existing state-of-the-art approaches, achieving strong performance even using only raw geometric features without relying on texture or appearance cues.
\end{itemize}

\section{Related Work}

Unsupervised 3D anomaly detection has gained increasing attention with the introduction of the MVTec 3D‑AD dataset \cite{bergmann2021mvtec}, which provides depth scans across ten object categories with annotated structural defects such as scratches, dents, and deformations. On the other hand, Real3D-AD \cite{liu2023real3d} is a benchmark that focuses on high-resolution point clouds, where each prototype contains forty thousand to millions of points with complete 360° geometric coverage. Unlike MVTec 3D-AD, which uses depth maps and voxelized inputs, Real3D-AD provides raw point-based 3D data with fine structural detail, enabling more precise anomaly localization while posing greater computational and generalization challenges. Existing methods can be grouped into the following categories. 

Reconstruction-based methods train models \cite{bergmann2021mvtec,zhou2024r3d,li2024towards,cheng2025mc3d,liang2025taming} such as autoencoders to reconstruct clean point clouds and identify anomalies by comparing input and output geometry. For example, the Voxel Autoencoder baseline in the MVTec 3D-AD benchmark encodes voxelized 3D input data into a latent space and reconstructs it through a decoder \cite{bergmann2021mvtec}. Anomalies are detected by evaluating voxel-wise reconstruction errors. However, due to the lossy compression and limited capacity of the autoencoder, the reconstructions tend to be blurry and imprecise, especially around object boundaries. As a result, subtle geometric defects are often oversmoothed and missed, leading to reduced localization accuracy and poor sensitivity to fine structural deviations. IMRNet\cite{li2024towards} adopts a self-supervised masked reconstruction strategy, where the network learns to recover missing regions in a point cloud, thereby capturing structural regularities of normal objects for effective anomaly detection.

Memory-based methods \cite{liu2023real3d,zhao2024pointcore,wang2023multimodal,zhu2024towards} store representative features from normal samples in a feature bank and detect anomalies by comparing test samples against this memory. For example, in Reg3D-AD \cite{liu2023real3d}, PatchCore is adapted from the 2D PatchCore framework, matches local descriptors such as FPFH or PointMAE features between test and reference regions. However, this matching process lacks spatial constraint, as features are retrieved globally across the entire object. This can lead to context-insensitive errors: test points that are geometrically abnormal may still be matched to unrelated but similar-looking regions elsewhere in the object, especially in the presence of repetitive structures or symmetry. This may lead to false positives or missed defects. To address this, PointCore \cite{zhao2024pointcore} proposes a joint local-global feature scheme that anchors the matching process with spatially meaningful coordinate references, thus enhancing contextual fidelity while reducing computational overhead.

Another line of work leverages view-projection-based methods, which transform 3D point clouds into 2D depth or image views to exploit the power of pretrained 2D networks. For instance, CPMF \cite{cao2024complementary} projects point clouds into multiple depth maps and fuses 2D semantic features with handcrafted 3D descriptors to enhance robustness and spatial awareness. Similarly, MVP \cite{cheng2024towards} formulates 3D anomaly detection as a zero-shot task by aligning multi-view depth projections with vision-language models, enabling generalization to unseen categories. These approaches benefit from strong 2D model priors and cross-modal flexibility, but may suffer from projection distortion and reduced point-wise localization accuracy. ISMP \cite{liang2025look} is a 3D anomaly detection method that captures internal spatial structure by slicing point clouds into pseudo-modal projections from multiple internal perspectives. These views enrich global representations beyond traditional external-only approaches. ISMP further enhances local features by combining PointMAE and FPFH-based patches, and applies Laplacian-based filtering to suppress noise and improve structural consistency.

Teacher–student frameworks adapt 2D distillation paradigms to 3D \cite{qin2023teacher, bergmann2023anomaly, rudolph2023asymmetric} by learning to replicate dense geometric descriptors from a pretrained teacher network \cite{bergmann2023anomaly}. AST \cite{rudolph2023asymmetric} improves detection by combining a normalizing flow teacher with a lightweight student, triggering stronger divergence for anomalies. While effective at object-level detection, these models struggle with spatial granularity. Classical methods like BTF \cite{horwitz2023back} surprisingly outperform deep learning baselines using handcrafted, rotation-invariant features. However, their limited adaptability constrains generalization.  Lightweight solutions like EasyNet \cite{chen2023easynet} prioritize deployment feasibility with compact architectures and fast inference but may compromise localization. Overall, while prior work has advanced detection accuracy, challenges remain in balancing fine-grained localization, generalization across shape variations, and computational efficiency. These gaps motivate our method, which leverages keypoint-guided clustering and registration to improve robustness and resolution in high-resolution 3D anomaly detection.

\section{Method}

\begin{figure}[tb] 
\centering 
\includegraphics[width=1.0\linewidth]{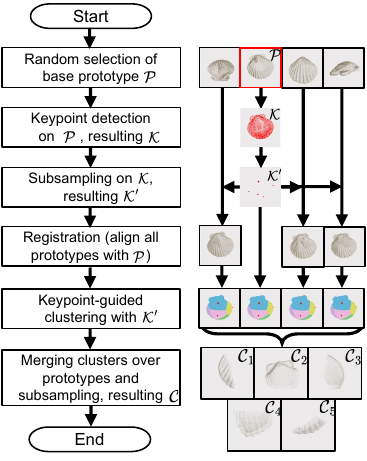} 
\caption{Flowchart of the preprocessing stage over prototypes.}
\label{fig3} 
\end{figure}

Our method consists of two main stages: a preprocessing stage that constructs a structured comparison reference from normal prototypes, and an inference stage that aligns each test sample with this reference for anomaly scoring. We denote point clouds as sets of 3D coordinates. Our method is built upon our previous conference work \cite{wang2025boosting} by introducing the key-point based point clustering.

\subsection{Preprocessing Stage}

The flowchart of the preprocessing stage is shown in Fig. \ref{fig3}. Given a set of $N$ normal prototypes $\{\mathcal{P}_1, \ldots, \mathcal{P}_N\}$, we first randomly select a base prototype $\mathcal{P}$ for alignment reference. Then,
\setlength{\leftmargini}{0pt}
\begin{enumerate}
\item \textbf{Keypoint Detection:} Apply a 3D keypoint detector to $\mathcal{P}$ to extract salient geometric locations:
\begin{equation}
    \mathcal{K} = \text{KeypointDetect}(\mathcal{P})
\end{equation}
The choice of keypoint detector is flexible and not restricted to a specific algorithm. In our experiments, we compare several classical keypoint detection methods, including Harris3D, Harris6D~\cite{harris1988combined}, and SIFT-3D~\cite{lowe2004distinctive}, to evaluate their impact on anomaly detection performance.

\item \textbf{Keypoint Subsampling:} Downsample $\mathcal{K}$ to obtain a fixed number of centroids for clustering:
\begin{equation}
    \mathcal{K}' = \text{Subsample}(\mathcal{K})
\end{equation}
In practice, various subsampling strategies can be applied. In our experiments, we evaluate several commonly used methods including uniform sampling, random sampling, and furthest point sampling, to investigate their influence on downstream performance.

\item \textbf{Prototype Registration:} Align all other prototypes to the base prototype $\mathcal{P}$ via rigid registration:
\begin{equation}
    \hat{\mathcal{P}}_i = \text{Register}(\mathcal{P}_i, \mathcal{P}), \quad \forall i \in \{1, \ldots, N\}, \mathcal{P}_i \neq \mathcal{P}
\end{equation}
To facilitate consistent structural comparison, all prototype point clouds are registered to a common coordinate system defined by a randomly selected anchor $\mathcal{P}$. Given a target prototype $\mathcal{P}_i$ and the anchor $\mathcal{P}$, we estimate a rigid transformation $\mathbf{T}_i$ that minimizes the alignment loss:
\begin{equation}
\mathbf{T}_i^* = \arg\min_{\mathbf{T}} \mathcal{L}(\mathbf{T} \cdot \mathcal{P}_i, \mathcal{P}).
\end{equation}
We follow the two-stage registration procedure proposed in \cite{liu2023real3d}: an initial transformation is obtained via global registration (RANSAC) using FPFH features \cite{rusu2009fast}, followed by local refinement using point-to-plane ICP \cite{rusinkiewicz2001efficient}. The global alignment minimizes a point-to-point Euclidean loss:
\begin{equation}
\mathcal{L}_{\text{p2p}}(\mathbf{T}) = \sum_{(i, j) \in {Corr}} \left\| \mathbf{T} \cdot \mathbf{p}_i^p - \mathbf{p}_j^a \right\|_2^2,
\end{equation}
where ${Corr}$ is the set of matched keypoint pairs obtained via feature correspondences. The ICP refinement minimizes a point-to-plane loss to account for surface orientation:
\begin{equation}
\mathcal{L}_{\text{p2pl}}(\mathbf{T}) = \sum_{(i, j) \in {Corr}} \left( \mathbf{n}_j^\top \left( \mathbf{T} \cdot \mathbf{p}_i^p - \mathbf{p}_j^a \right) \right)^2,
\end{equation}
where $\mathbf{n}_j$ is the normal at point $\mathbf{p}_j^a$. This alignment strategy ensures that all prototypes share a unified spatial reference, enabling reliable local feature comparison in inference stage.

\item \textbf{Keypoint-Guided Clustering:} For each registered prototype $\hat{\mathcal{P}}_i$, assign each point $p \in \hat{\mathcal{P}}_i$ to its nearest keypoint in $\mathcal{K}'$. The resulting clusters are denoted as $\mathcal{C}_{ij}$, where $j$ indexes the $j$-th keypoint:
\begin{equation}
\mathcal{C}_{ij} = \left\{ p \in \hat{\mathcal{P}}_i \;\middle|\; j = \arg\min_{k = 1, \ldots, |\mathcal{K}'|} \| p - \mathcal{K}'_k \|_2 \right\}
\end{equation}
    
\item \textbf{Cluster Merging and Subsampling:} For each cluster index $j$, merge the corresponding clusters $\mathcal{C}_{ij}$ from all $N$ prototypes and subsample to a uniform size:
\begin{equation}
\mathcal{C}_j = \text{Subsample} \left( \bigcup_{i=1}^{N} \mathcal{C}_{ij} \right), \quad \forall j = 1, \ldots, |\mathcal{K}'|
\end{equation}
\end{enumerate}

The final result of preprocessing is the set of reference clusters $\mathcal{C} = \{\mathcal{C}_{j}\}, \forall j = 1, \ldots, |\mathcal{K}'|$

\begin{figure}[tb] 
\centering 
\includegraphics[width=1.0\linewidth]{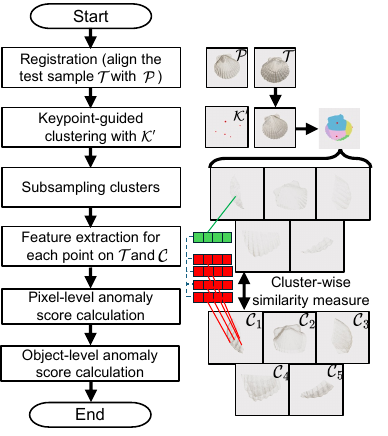} 
\caption{Flowchart of the inference stage over test data. }
\label{fig4} 
\end{figure}

\subsection{Inference Stage}

The flowchart of the inference stage is shown in Fig.~\ref{fig4}. Given a test sample $\mathcal{T}$, we follow the same clustering strategy established in the preprocessing stage. The test sample is first aligned to the base prototype $\mathcal{P}$ and clustered using the keypoint set $\mathcal{K}'$. Based on each test cluster $\mathcal{C}^{T}_j$ corresponding to reference cluster $\mathcal{C}_j$, we compute point-level and object-level anomaly scores as follows.

\begin{enumerate}
    \item \textbf{Registration and Clustering:} Align the test sample to the base prototype and assign each point $t \in \hat{\mathcal{T}}$ to its nearest keypoint $\mathcal{K}'_j$ to form $\mathcal{C}^{T}_j$:
    \begin{equation}
        \mathcal{C}^{T}_j = \left\{ t \in \hat{\mathcal{T}} \;\middle|\; j = \arg\min_{k = 1, \ldots, |\mathcal{K}'|} \| t - \mathcal{K}'_k \|_2 \right\}
    \end{equation}

    \item \textbf{Point-Level Anomaly Score:} For each point $t^j_i \in \mathcal{C}^{T}_j$, we compute the anomaly score with respect to the pooled reference cluster $\mathcal{C}_j = \bigcup_{i=1}^{N} \mathcal{C}_{ij}$. The anomaly score $s^\ast_{ij}$ for $t^j_i$ is defined as:
    \begin{equation}
        s^\ast_{ij} = \min_{q \in \mathcal{C}_j} \| t^j_i - q \|_2
    \end{equation}
\begin{enumerate}
    \item \textbf{Reweighting:} Inspired by RegAD \cite{liu2023real3d}, to account for local geometric consistency, we compute a reweighting term $w_{ij}$ for each point $t^j_i$ using its $K$ nearest neighbors in $\mathcal{C}_j$:
\begin{equation}
\begin{aligned}
    w_{ij} &= 1 - \frac{\exp\left( \frac{s^\ast_{ij}}{\vert \mathcal{C}^{T}_j \vert} \right)}{\sum_{k=1}^{K} \exp\left( \frac{\Vert t^j_i - n_k \Vert_2}{\vert \mathcal{C}^{T}_j \vert} \right)} \\
    &\text{where } \{ n_k \}_{k=1}^{K} = \text{topK}(t^j_i, \mathcal{C}_j)
\end{aligned}
\end{equation}

\item \textbf{Weighted Point Score:} The final anomaly score for $t^j_i$ is given by:
\begin{equation}
    s_{ij} = w_{ij} \cdot s^\ast_{ij}
\end{equation}

\item \textbf{Point-Level Interpolation:} To upsample point-wise anomaly scores to the full resolution of $\mathcal{T}$, we use $K$-nearest neighbor interpolation. For each point $t_i^{j}$ ($i$-th point in the $j$-th cluster) in the original test point cloud:
\begin{equation}
s^{pt}_{ij} = \frac{1}{|\mathcal{N}_i^j|} \sum_{m \in \mathcal{N}_i^j} s_{mj}, \quad \mathcal{N}_i^j = \text{KNN}(t_i^{j}, \mathcal{C}^{T}_j)
\end{equation}
\end{enumerate}
    \item \textbf{Object-Level Anomaly Score:} The object-level anomaly score for test sample $\mathcal{T}$ is defined conservatively as the maximum of the minimum distances across all clusters:
    \begin{equation}
        s^{obj} = \max_{j=1,\ldots,|\mathcal{K}'|} \min_{i=1,\ldots,|\mathcal{C}^{T}_j|} s_{ij}^{pt}
    \end{equation}

    \item \textbf{Multi-Feature Fusion (Optional):} If multiple feature types (e.g., raw coordinates and FPFH) are used, their scores can be fused as:
    \begin{equation}
        s^{pt}_{ij} = \lambda s^{raw}_{ij} + (1 - \lambda) s^{FPFH}_{ij}
    \end{equation}
\end{enumerate}

\section{Experiment}
\subsection{Dataset and Evaluation Metrics}

We evaluate our method on the Real3D-AD dataset~\cite{liu2023real3d}, a high-resolution benchmark designed for industrial point cloud anomaly detection. It consists of 1,254 samples from 12 object categories, each including four clean prototypes and over 100 test samples with annotated normal and anomalous regions. For performance evaluation, we follow the standard protocol in~\cite{liang2025look} and report both object-level and point-level detection metrics. For fair comparison, we follow the recent work \cite{liang2025look} to evaluate our method. Except for the results of our method, the results of all the comparative methods replicate the numerical results provided in \cite{liang2025look}.

\noindent\textbf{Object-level AUROC (O-AUROC)} evaluates the ability of a model to distinguish between normal and anomalous 3D objects using scalar anomaly scores. It is computed using the standard Area Under the Receiver Operating Characteristic curve:
\begin{equation}
\text{O-AUROC} = \text{AUROC}( \{y^{\text{obj}}_i\}, \{s^{\text{obj}}_i\} ),
\end{equation}
where $y^{\text{obj}}_i \in \{0,1\}$ denotes the ground truth label (normal or anomalous) and $s^{\text{obj}}_i$ the predicted anomaly score for the $i$-th object.

\noindent\textbf{Point-level AUROC (P-AUROC)} assesses the localization performance by comparing point-wise predicted scores against binary ground truth labels for each point:
\begin{equation}
\text{P-AUROC} = \text{AUROC}( \{y^{\text{pt}}_j\}, \{s^{\text{pt}}_j\} ),
\end{equation}
where $y^{\text{pt}}_j \in \{0,1\}$ and $s^{\text{pt}}_j$ represent the label and anomaly score for the $j$-th point, respectively.

\begin{table*}[tb]
\caption{O-AUROC ($\uparrow$) over 12 categories in the Real3D-AD dataset. The highest and second-highest scores are in bold and underlined, respectively. Our 3DKeyAD achieved the top average performance. US/RS/FS: Uniform/Random/Furthest Sampling.}
    \label{tab:1}
     \begin{tabularx}{\linewidth}{lYYYYYYYYYYYY|Y}
        \hline
        \textbf{Method} & \tiny \textbf{Airplane} & \tiny \textbf{Car} & \tiny \textbf{Candybar} & \tiny \textbf{Chicken} & \tiny \textbf{Diamond} & \tiny \textbf{Duck} & \tiny \textbf{Fish} & \tiny \textbf{Gemstone} & \tiny \textbf{Seahorse} & \tiny \textbf{Shell} & \tiny \textbf{Starfish} & \tiny \textbf{Toffees} & \tiny \textbf{Mean} \\
        \hline
        BTF(Raw) \cite{horwitz2023back} {\textit{CVPR2023}} & 0.730 & 0.647 & 0.539 & 0.789 & 0.707 & 0.691 & 0.602 & \textbf{{0.686}} & 0.596 & 0.396 & 0.530 & 0.703 & 0.635 \\ 
        BTF(FPFH) \cite{horwitz2023back} {\textit{CVPR2023}} & 0.520 & 0.560 & 0.630 & 0.432 & 0.545 & \textbf{{0.784}} & 0.549 & 0.648 & 0.779 & 0.754 & 0.575 & 0.462 & 0.603 \\ 
        M3DM \cite{wang2023multimodal} {\textit{CVPR2023}} & 0.434 & 0.541 & 0.552 & 0.683 & 0.602 & 0.433 & 0.540 & 0.644 & 0.495 & 0.694 & 0.551 & 0.450 & 0.552 \\ 
        PatchCore(FPFH) \cite{roth2022towards} {\textit{CVPR2022}} & \textbf{{0.882}} & 0.590 & 0.541 & \underline{{0.837}} & 0.574 & 0.546 & 0.675 & 0.370 & 0.505 & 0.589 & 0.441 & 0.565 & 0.593 \\ 
        PatchCore(PointMAE) \cite{roth2022towards} {\textit{CVPR2022}} & 0.726 & 0.498 & 0.663 & 0.827 & 0.783 & 0.489 & 0.630 & 0.374 & 0.539 & 0.501 & 0.519 & 0.585 & 0.594 \\ 
        CPMF \cite{cao2024complementary} {\textit{Pattern Recognit.2024}} & 0.701 & 0.551 & 0.552 & 0.504 & 0.523 & 0.582 & 0.558 & 0.589 & 0.729 & 0.653 & \textbf{{0.700}} & 0.390 & 0.586 \\ 
        RegAD \cite{liu2023real3d} {\textit{NeurIPS2023}} & 0.716 & 0.697 & 0.685 & \textbf{{0.852}} & 0.900 & 0.584 & 0.915 & 0.417 & 0.762 & 0.583 & 0.506 & 0.827 & 0.704 \\ 
        IMRNet \cite{li2024towards} {\textit{CVPR2024}} & 0.762 & 0.711 & 0.755 & 0.780 & 0.905 & 0.517 & 0.880 & \underline{{0.674}} & 0.604 & 0.665 & \underline{{0.674}} & 0.774 & 0.725 \\ 
        ISMP \cite{liang2025look} {\textit{AAAI2025}} & \underline{{0.858}} & 0.731 & {{0.852}} & 0.714 & 0.948 & 0.712 & {{0.945}} & 0.468 & 0.729 & 0.623 & 0.660 & 0.842 & 0.767 \\ \hline
3DKeyAD(Raw+FS) & 0.601 & {{0.774}} & 0.788 & 0.804 & \textbf{{0.962}} & 0.738 & {{0.960}} & 0.488 & {{\underline{0.994}}} & {{0.782}} & 0.596 & {{0.902}} & {{0.782}} \\
3DKeyAD(Raw+ISS+US)  & 0.609 & 0.806 & 0.629 & 0.770 & 0.929 & \underline{0.775} & 0.942 & 0.446 & \underline{0.994} & 0.839 & 0.566 & 0.885 & 0.766 \\
3DKeyAD(Raw+ISS+RS)  & 0.656 & 0.806 & 0.770 & 0.727 & 0.955 & 0.709 & 0.950 & 0.530 & \underline{0.994} & \textbf{0.865} & 0.582 & \underline{0.905} & 0.787 \\
3DKeyAD(Raw+ISS+FS)  & 0.616 & 0.806 & 0.783 & 0.803 & 0.940 & 0.760 & 0.960 & 0.502 & {0.988} & 0.833 & 0.590 & \textbf{0.925} & \underline{0.792} \\
3DKeyAD(Raw+Harris3D+FS) & 0.605 & \underline{0.819} & 0.796 & 0.800 & 0.950 & 0.745 & 0.958 & 0.501 & 0.980 & 0.804 & 0.593 & 0.891 & 0.790 \\
3DKeyAD(Raw+Harris6D+FS) & 0.577 & 0.802 & 0.779 & 0.769 & 0.949 & 0.696 & 0.958 & 0.466 & 0.980 & 0.786 & 0.630 & 0.858 & 0.771 \\
3DKeyAD(Raw+SIFT-3D+FS)  & 0.662 & \textbf{0.822} & 0.668 & 0.811 & 0.937 & 0.731 & \textbf{0.972} & 0.462 & \textbf{1.000} & \underline{0.847} & 0.584 & 0.859 & 0.780 \\
3DKeyAD(FPFH+ISS+FS) & 0.711 & 0.648 & \textbf{0.883} & 0.584 & 0.894 & 0.696 & 0.842 & 0.620 & 0.568 & 0.558 & 0.631 & 0.488 & 0.680 \\
3DKeyAD(PointMAE+ISS+FS) & 0.694 & 0.589 & 0.558 & 0.572 & 0.812 & 0.560 & 0.618 & 0.412 & 0.663 & 0.526 & 0.442 & 0.636 & 0.590 \\ 
3DKeyAD(Raw+PointMAE+ISS+FS) & 0.616 & 0.804 & 0.783 & 0.804 & 0.941 & 0.762 & 0.958 & 0.502 & {0.988} & 0.833 & 0.590 & \textbf{0.925} & \underline{0.792} \\ 
3DKeyAD(Raw+FPFH+ISS+FS) & 0.652 & 0.812 & \underline{0.875} & 0.794 & \underline{0.959} & 0.764 & \underline{0.963} & 0.520 & 0.982 & 0.799 & 0.592 & 0.904 & \textbf{0.801} \\ \hline
    \end{tabularx}
\end{table*}

\begin{table*}[tb]
    \centering
    \caption{P-AUROC ($\uparrow$) over 12 categories in the Real3D-AD dataset. The highest and second-highest scores are in bold and underlined, respectively. Our 3DKeyAD achieved the top average performance. US/RS/FS: Uniform/Random/Furthest Sampling.}
    \label{tab:2}
    \begin{tabularx}{\linewidth}{lYYYYYYYYYYYY|Y}
        \hline
        \textbf{Method} & \tiny \textbf{Airplane} & \tiny \textbf{Car} & \tiny \textbf{Candybar} & \tiny \textbf{Chicken} & \tiny \textbf{Diamond} & \tiny \textbf{Duck} & \tiny \textbf{Fish} & \tiny \textbf{Gemstone} & \tiny \textbf{Seahorse} & \tiny \textbf{Shell} & \tiny \textbf{Starfish} & \tiny \textbf{Toffees} & \tiny \textbf{Mean} \\
        \hline
        BTF(Raw) \cite{horwitz2023back} {\textit{CVPR2023}} & 0.564 & 0.647 & 0.735 & 0.609 & 0.563 & 0.601 & 0.514 & 0.597 & 0.520 & 0.489 & 0.392 & 0.623 & 0.571 \\
        BTF(FPFH) \cite{horwitz2023back} {\textit{CVPR2023}} & 0.738 & 0.708 & 0.864 & 0.735 & 0.882 & 0.875 & 0.709 & 0.891 & 0.512 & 0.571 & 0.501 & 0.815 & 0.733 \\ 
        M3DM \cite{wang2023multimodal} {\textit{CVPR2023}} & 0.547 & 0.602 & 0.679 & 0.678 & 0.608 & 0.667 & 0.606 & 0.674 & 0.560 & 0.738 & 0.532 & 0.682 & 0.631 \\ 
        PatchCore(FPFH) \cite{roth2022towards} {\textit{CVPR2022}} & 0.562 & 0.754 & 0.780 & 0.429 & 0.828 & 0.264 & 0.829 & \underline{{0.910}} & 0.739 & 0.739 & 0.606 & 0.747 & 0.682 \\ 
        PatchCore(PointMAE) \cite{roth2022towards} {\textit{CVPR2022}} & 0.569 & 0.609 & 0.627 & 0.729 & 0.718 & 0.528 & 0.717 & 0.444 & 0.633 & 0.709 & 0.580 & 0.580 & 0.620 \\
        RegAD \cite{liu2023real3d} {\textit{NeurIPS2023}} & 0.631 & 0.718 & 0.724 & 0.676 & 0.835 & 0.503 & 0.826 & 0.545 & 0.817 & 0.811 & 0.617 & 0.759 & 0.705 \\ 
        ISMP \cite{liang2025look} {\textit{AAAI2025}} & \underline{{0.753}} & 0.836 & 0.907 & 0.798 & \underline{{0.926}} & {{0.876}} & 0.886 & 0.857 & 0.813 & 0.839 & 0.641 & \underline{{0.895}} & \underline{{0.836}} \\ \hline
3DKeyAD(Raw+FS) & 0.734 & \underline{{0.866}} & 0.830 & 0.851 & 0.884 & 0.778 & 0.879 & 0.580 & {{0.901}} & \underline{{0.872}} & {{0.679}} & 0.868 & 0.810 \\
3DKeyAD(Raw+ISS+US)  & 0.726 & 0.841 & 0.807 & 0.841 & 0.885 & 0.817 & 0.876 & 0.622 & 0.899 & 0.869 & 0.710 & 0.876 & 0.814 \\
3DKeyAD(Raw+ISS+RS)  & 0.773 & 0.849 & 0.836 & 0.831 & 0.882 & 0.768 & 0.875 & 0.650 & \underline{0.910} & 0.866 & 0.715 & 0.866 & 0.818 \\
3DKeyAD(Raw+ISS+FS)  & 0.745 & 0.855 & 0.818 & 0.834 & 0.885 & 0.781 & 0.876 & 0.593 & 0.904 & 0.868 & 0.711 & 0.867 & 0.811 \\
3DKeyAD(Raw+Harris3D+FS)  & 0.716 & 0.846 & 0.831 & 0.841 & 0.877 & 0.789 & 0.868 & 0.626 & \underline{0.910} & 0.866 & \underline{0.716} & 0.870 & 0.813 \\
3DKeyAD(Raw+Harris6D+FS) & 0.734 & 0.847 & 0.834 & 0.855 & 0.876 & 0.785 & 0.870 & 0.562 & \underline{0.910} & 0.865 & 0.714 & 0.882 & 0.811 \\
3DKeyAD(Raw+SIFT-3D+FS)  & 0.701 & 0.855 & 0.800 & 0.850 & 0.884 & 0.793 & 0.876 & 0.651 & 0.909 & 0.865 & \textbf{0.721} & 0.854 & 0.813 \\ 
3DKeyAD(FPFH+ISS+FS) & 0.748 & 0.725 & \underline{0.922} & 0.703 & 0.912 & \textbf{0.894} & 0.788 & \textbf{0.932} & 0.548 & 0.639 & 0.534 & 0.845 & 0.770 \\
3DKeyAD(PointMAE+ISS+FS) & 0.573 & 0.665 & 0.644 & 0.707 & 0.712 & 0.565 & 0.726 & 0.457 & 0.658 & 0.756 & 0.598 & 0.747 & 0.650 \\ 
3DKeyAD(Raw+PointMAE+ISS+FS) & 0.749 & \textbf{0.870} & 0.825 & \underline{0.858} & 0.897 & 0.785 & \textbf{0.902} & 0.582 & \textbf{0.913} & \textbf{0.900} & 0.714 & 0.886 & 0.820 \\ 
3DKeyAD(Raw+FPFH+ISS+FS) & \textbf{0.788} & 0.838 & \textbf{0.935} & \textbf{0.867} & \textbf{0.948} & \underline{0.884} & \underline{0.898} & 0.838 & 0.863 & 0.861 & 0.701 & \textbf{0.913} & \textbf{0.861} \\ \hline
    \end{tabularx}
\end{table*}

\begin{figure*}[tb] 
\centering 
\includegraphics[width=1.0\linewidth]{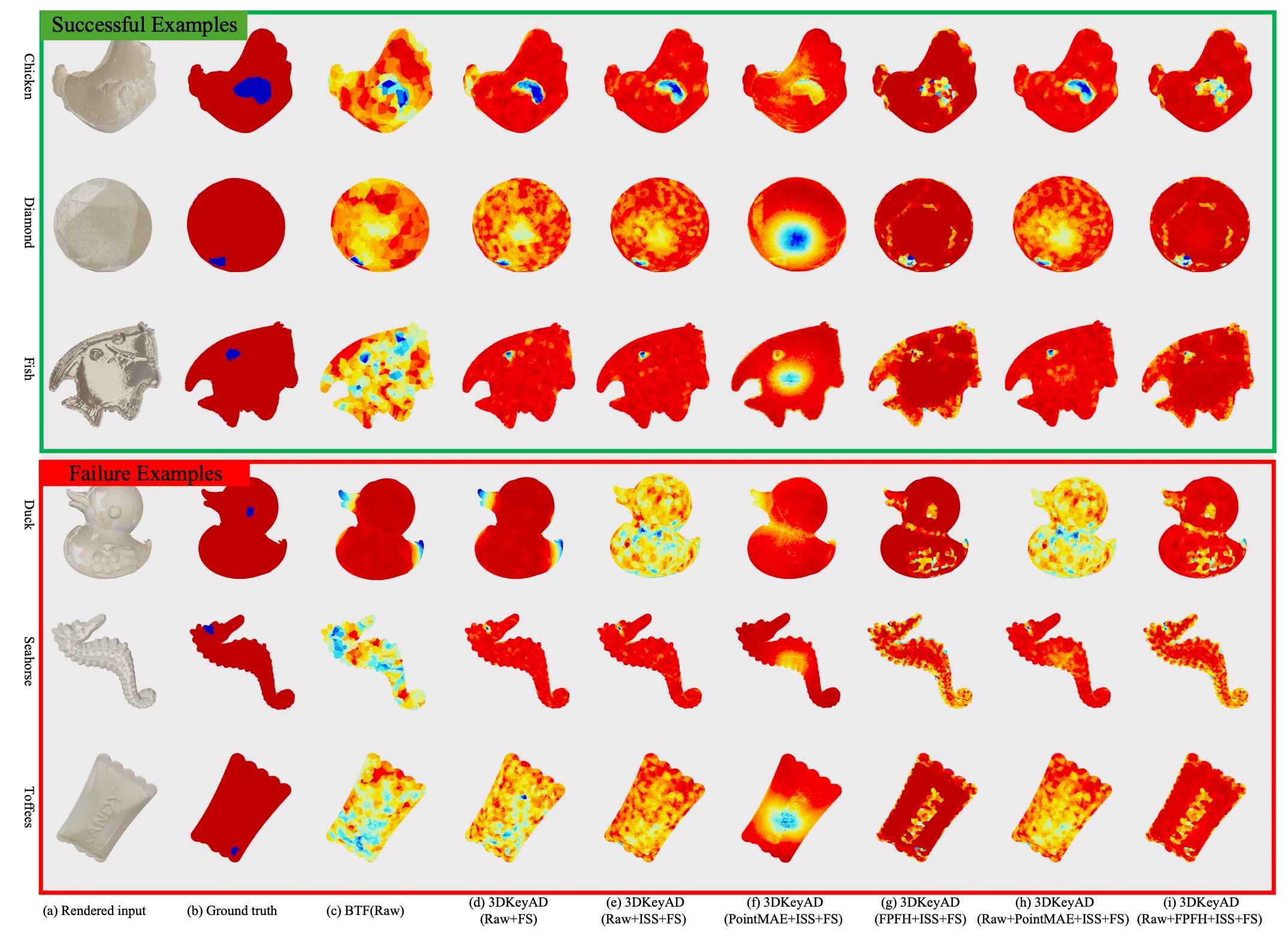} 
\caption{Qualitative results on Real3D-AD. Top: successful cases with accurate localization. Bottom: failure cases due to subtle defects (\textit{Duck}), complex geometry (\textit{Seahorse}), or fine normal textures (\textit{Toffees}). Columns show input, ground truth, and predictions from different 3DKeyAD variants.}
\label{fig5} 
\end{figure*}

\subsection{Implementation Details}

To ensure fair comparisons across different feature types, we applied a consistent subsampling strategy throughout all experiments. Each prototype was downsampled to only 20,000 points, and each test sample to only 1,000 points. The number of subsampled keypoints was fixed at 5. For anomaly score interpolation, we set the number of nearest neighbors to 3 and used a reweighting factor $\lambda = 0.01$.
As shown in Table~\ref{tab:1} and Table~\ref{tab:2}, we construct multiple ablation variants to investigate the effects of keypoint detection strategies and feature types. In the baseline variant 3DKeyAD (Raw+FS), keypoint detection is omitted. Instead, furthest point sampling (FPS) is directly applied to the full point cloud, and the sampled points are used as cluster centroids. Feature similarity is computed using raw 3D coordinates.
To evaluate the impact of keypoint selection, we adopt the Intrinsic Shape Signatures (ISS) detector~\cite{zhong2009intrinsic} to extract geometrically salient points. We then explore three subsampling methods on the detected keypoints: uniform sampling (US), random sampling (RS), and furthest point sampling (FS), resulting in three variants: 3DKeyAD (Raw+ISS+US), (Raw+ISS+RS), and (Raw+ISS+FS).
We further experiment with alternative keypoint detectors, including Harris3D, Harris6D~\cite{harris1988combined}, and SIFT-3D~\cite{lowe2004distinctive}, to assess the sensitivity of anomaly detection performance to different detection strategies.
Finally, we investigate the effect of feature representation on cluster-level comparison. In addition to raw coordinates, we evaluate handcrafted descriptors such as FPFH, pretrained features from PointMAE, and their combinations, resulting in variants such as Raw+FPFH and Raw+PointMAE.
All experiments were conducted on a desktop with an Intel Core i9-12900K CPU, 32\,GB RAM, and an NVIDIA RTX 3090 GPU with 24\,GB VRAM.

\subsection{Quantitative Results}

We report the object-level (O-AUROC) and point-level (P-AUROC) anomaly detection performance across 12 categories of the Real3D-AD dataset in Table~\ref{tab:1} and Table~\ref{tab:2}, respectively. Our proposed framework, 3DKeyAD, consistently outperforms existing state-of-the-art methods on both metrics.

On the object-level task, 3DKeyAD (Raw+FPFH+ISS +FS) achieves the best average score of 0.801, surpassing ISMP (0.767)\cite{liang2025look} and RegAD (0.704)\cite{liu2023real3d}. Even the baseline variant 3DKeyAD (Raw+FS), which uses no keypoint detection, outperforms memory bank based approaches such as M3DM~\cite{wang2023multimodal} and PatchCore~\cite{roth2022towards}. On the point-level localization task, 3DKeyAD (Raw+FPFH+ISS+FS) again achieves the highest mean score of 0.861, outperforming ISMP (0.836) and RegAD (0.705). The combination of raw and FPFH features significantly enhances localization precision, and the use of ISS keypoints with furthest sampling further contributes to stability across object types. Among all keypoint variants, Harris3D, Harris6D, and SIFT-3D also yield competitive results, confirming the framework’s flexibility with different detectors. Meanwhile, 3DKeyAD (Raw+PointMAE+ISS+FS) performs best in categories like Car and Seahorse, showing that deep learned features can complement handcrafted ones. However, using PointMAE alone performs poorly, which aligns with observations in \cite{horwitz2023back}. Deep-learned features such as those from PointMAE often perform suboptimally in 3D anomaly detection due to several factors. First, they are typically sensitive to rotation and may lack the robustness needed for arbitrary object orientations. Second, pretrained models may introduce domain gaps, as their learned representations are not specifically tuned to capture fine-grained geometric deviations common in industrial defects. Lastly, deep features tend to overgeneralize, smoothing out subtle structural cues that are critical for accurate anomaly localization. In contrast, classical descriptors like FPFH retain local geometric details and offer stronger rotation invariance, making them more effective for this task. Overall, the improvement in both object-level and point-level AUROC metrics demonstrate the effectiveness of our keypoint-guided clustering and multi-prototype alignment strategy for high-resolution 3D anomaly detection.

\subsection{Qualitative Results}
As shown in Fig. \ref{fig5}, we present visual comparisons of anomaly localization results on both successful and failure cases across various categories in the Real3D-AD dataset. Each row corresponds to a specific object class, and each column shows the prediction from a different method.

For categories such as \textit{Chicken}, \textit{Diamond}, and \textit{Fish}, our proposed 3DKeyAD variants (particularly Raw+ISS+FS and Raw+FPFH+ISS+FS) produce anomaly maps that closely match the ground-truth annotations. These methods accurately highlight small and scattered defects while suppressing false positives in normal regions. In contrast, BTF(Raw) tends to produce noisy or overly diffused predictions, missing fine-grained anomalies or triggering irrelevant regions. Notably, the use of handcrafted features (FPFH) or hybrid feature combinations improves the spatial sharpness and reliability of detection.

Despite the overall effectiveness of our method, certain failure cases reveal key limitations under challenging conditions. For example, in categories like \textit{Duck}, anomalies are extremely small and seamlessly blended into the normal surface, making them difficult to distinguish even by human observers, and both raw and learned features fail to capture such subtle deviations. In structurally complex objects such as \textit{Seahorse}, which exhibit significant geometric variation and intricate shapes, local feature descriptors like FPFH or deep embeddings from PointMAE may produce high false positives, as natural intra-class diversity is often mistaken for structural defects. Additionally, in categories like \textit{Toffees}, where normal samples contain fine-grained embossed patterns (e.g., text like “candy”), feature-based anomaly detection may incorrectly flag these semantic but structurally irregular patterns as anomalies, due to their uniqueness. These observations suggest that while keypoint-guided clustering improves robustness, feature-based detection still struggles in cases of subtle anomalies, complex geometry, or high-frequency normal patterns.

\section{Conclusion}
In this paper, we presented a registration-based framework for high-resolution 3D anomaly detection. Our method leverages multi-prototype alignment to ensure geometric consistency across samples, and introduces a keypoint-guided clustering strategy that defines semantically meaningful comparison regions for local anomaly scoring. This design effectively balances the need for structural flexibility and computational efficiency, enabling precise object-level and point-level detection. We validated our approach on the Real3D-AD benchmark, which provides dense, texture-free point clouds with only 3D coordinates. The dataset poses significant challenges due to its resolution and intra-class variability, yet our method consistently outperformed existing baselines in both detection accuracy and localization precision. Notably, our framework operates solely on geometric information, without relying on appearance cues or external supervision. Future work includes extending our method to dynamic or time-varying 3D data, incorporating lightweight learning-based refinement, and exploring integration with downstream tasks such as defect classification or repair planning.
\bibliographystyle{ieicetr}
\bibliography{egbib}
\end{document}